# Convolutional Autoencoder for Blind Hyperspectral Image Unmixing

Yasiru Ranasinghe
*School of Engineering*
*Sri Lanka Technological Campus*
Padukka 10500, Sri Lanka
yasirur@sltc.ac.lk

Sanjaya Herath
*Department of Electrical and Electronic Engineering*
*University of Peradeniya*
Peradeniya 20400, Sri Lanka
sanjaya.h@eng.pdn.ac.lk

Kavinga Weerasooriya
*Department of Electrical and Electronic Engineering*
*University of Peradeniya*
Peradeniya 20400, Sri Lanka
kavingaweerasooriya@eng.pdn.ac.lk

Mevan Ekanayake
*Office of Research and Innovation Services*
*Sri Lanka Technological Campus*
Padukka 10500, Sri Lanka
mevane@sltc.ac.lk

Roshan Godaliyadda
*Department of Electrical and Electronic Engineering*
*University of Peradeniya*
Peradeniya 20400, Sri Lanka
roshangodd@ee.pdn.ac.lk

Parakrama Ekanayake
*Department of Electrical and Electronic Engineering*
*University of Peradeniya*
Peradeniya 20400, Sri Lanka
mpb.ekanayake@ee.pdn.ac.lk

Vijitha Herath
*Department of Electrical and Electronic Engineering*
*University of Peradeniya*
Peradeniya 20400, Sri Lanka
vijitha@eng.pdn.ac.lk

*Abstract* — In the remote sensing context spectral unmixing is a technique to decompose a mixed pixel into two fundamental representatives: endmembers and abundances. In this paper, a novel architecture is proposed to perform blind unmixing on hyperspectral images. The proposed architecture consists of convolutional layers followed by an autoencoder. The encoder transforms the feature space produced through convolutional layers to a latent space representation. Then, from these latent characteristics the decoder reconstructs the roll-out image of the monochrome image which is at the input of the architecture; and each single-band image is fed sequentially. Experimental results on real hyperspectral data concludes that the proposed algorithm outperforms existing unmixing methods at abundance estimation and generates competitive results for endmember extraction with RMSE and SAD as the metrics, respectively.

*Keywords—Convolutional autoencoder (CAE), Hyperspectral unmixing, Convolutional neural networks (CNN), Deep learning, Endmember identification*

## I. INTRODUCTION

Hyperspectral image (HSI) data has been extensively used in remote sensing as it provides a glut of spectral information to analyze and classify materials[1]–[4] in an area. Though, the respective sensory methods often offer higher spectral resolution, the same cannot be told regarding their spatial resolution. Consequently, the measured reflectance is a mixture of spectral signatures of the underlying substances. Hence, a myriad of effort has gone into developing hyperspectral unmixing (HU) methods. In general, HU aims to decompose an amalgam of measured reflectance into the materials' spectral signatures, known as endmember signatures; and to estimate the fractional contribution by each endmember to the mixture, known as abundances.

The algorithms proposed for the HU problem can be categorized as statistical[5] and geometrical algorithms[6]. The former class of algorithms makes use of statistical measures to scrutinize a mixed pixel, whereas the latter group of algorithms exploits the Euclidean representation of a pixel in $\mathbb{R}^n$, where $n$ is the number of spectral bands of the HSI. Further, blind spectral unmixing has been profoundly performed through independent component analysis[7] with a variety of objective functions such as uncorrelation, kurtosis, entropy, etc. On the other hand, the same task has been performed using graph signal processing[8] incorporating nonnegative matrix factorization and different regularizations. It is noteworthy that there are physics-based solutions[9] as well, for the HU problem such as kernel methods, simulations, radiative transfer models, etc.

In the recent decade or so, data driven techniques that are encouraged by deep learning (DL) and neural networks (NN) have also proliferated in the field of HU. Advancements in computer hardware and software has facilitated the utilization of large spectral libraries to train DL based architectures[10] for unmixing. On the other hand, NN architectures have been developed to conduct blind source extraction[11] to produce endmembers and fractional abundance maps that require relatively less computational power. The inherent variability in the spectral information[12] of the HSI data is the widely used criterion for HU. Indeed, this has engendered the realization of traditional geometrical algorithms using standard NN architectures, specifically with autoencoder architectures.

In this paper, a convolutional autoencoder (CAE) architecture is proposed for the HU problem. The autoencoder is designed to perform blind unmixing on HSI data, given the number of endmembers to be extracted. It has been established that convolutional neural networks (CNN) are superior at detecting local features[13] of an image set. Owing to this aforementioned potential of CNN, the proposed method captures spatial distributions in an image through convolutional filters which are then used to parameterize innate spectral characteristics. The performance of the autoencoder was evaluated on the basis of endmember extraction and abundance estimation. For this spectral angle distance (SAD) and root mean square error (RMSE) were used as the numerical metrics, respectively. The autoencoder was trained with mean squared error (MSE) as the loss function to reduce the computational complexity.

The paper is organized as follows. In Section 2, the proposed method is delineated. In Section 3, experimental results are provided for available remote sensing datasets. Finally, in Section 4, conclusions are drawn.

## II. PROPOSED METHOD

In this section, the first subsection is dedicated to discuss the mathematical foundation of the autoencoder architecture for the endmember extraction and the abundance estimation. In the following section, the architecture of the CAE will be described with the specifications of the layers as given in Table I and hyper parameters used for the best-case scenario.

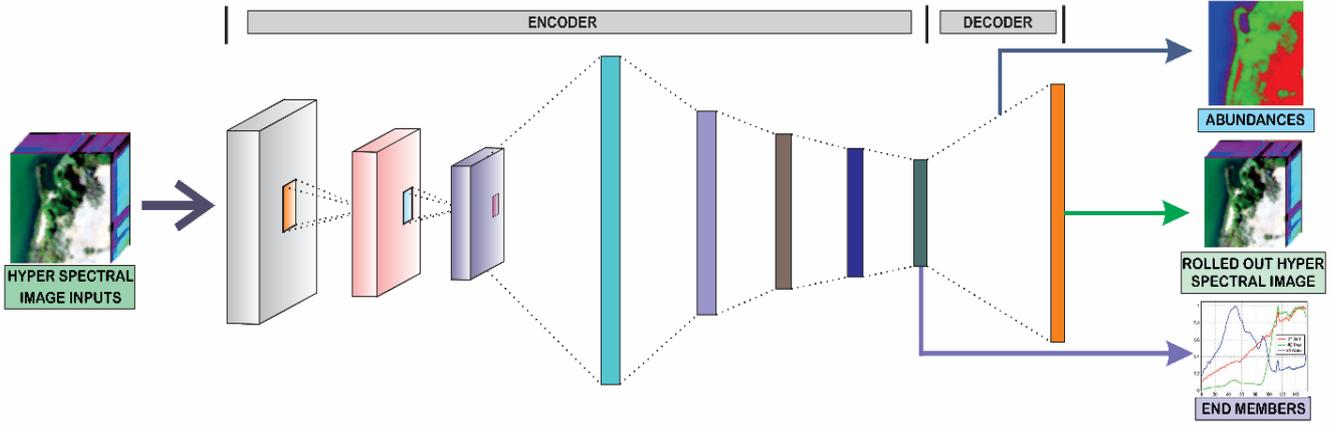

Fig. 1. Schematic of the convolutional autoencoder.

*A. Mathematical Foundation of the Architecture*

Given an HSI data matrix $X$, the objective is to decompose it into two matrices $A$ and $S$ as in equation 1. The matrix $A$ is called the endmember matrix and contains the spectral signatures of the endmembers, and the matrix $S$ is called the abundance matrix which contains the fractional contribution to each pixel. All the elements in $A, S$ and $X$ should be nonnegative.

$$X_{m \times n} = S_{m \times r} \times A_{r \times n} \quad (1)$$

Here, $n, m$, and $r$ are the number of spectral bands, the pixel resolution of the HSI and the number of endmembers to be extracted, respectively. The CAE attempts to reproduce the input given to it at its output. The autoencoder structure consists of two main sections. They are, namely, the encoder and the decoder. The encoder will transform the given input to a latent space, and the decoder attempts to reconstruct the input from the latent space.

Consider, the monochrome image corresponding to the $b$ spectral band, denoted by $x_b$ fed as the input to the convolutional autoencoder with $L$ layers. Then the output $\hat{x}_b$ is the reconstruction of the roll-out image of that input. If a rectified linear unit (ReLU) or a linear function is used as the activation for the output layer, the transformation performed at the output layer can be written as,

$$\hat{x}_b = W^{(L)} a^{(L-1)} \quad (2)$$

where, $a^{(L-1)}$ has the activations from the $(L-1)^{th}$ layer and $W^{(L)}$ are the weights of the output layer. If the $(L-1)^{th}$ layer has $r$ neurons, then the reconstructed HSI $\hat{X}$ can be written as,

$$\hat{X} = [\hat{x}_1, \hat{x}_2, \ldots, \hat{x}_b, \ldots, \hat{x}_n] \quad (3)$$

$$\hat{X} = W^{(L)} [a^{(L-1)(1)}, a^{(L-1)(2)}, \ldots, a^{(L-1)(m)}] \quad (4)$$

$$\hat{X} = W^{(L)}_{m \times r} \times H^{(L-1)}_{r \times n} \quad (5)$$

where, $H^{(L-1)}_{r \times n}$ is the matrix constructed using activations of the $(L-1)^{th}$ layer for all pixels. When comparing this with equation 1, it is evident that the endmember matrix $A_{r \times n}$ corresponds to the matrix formed by the activations of $(L-1)^{th}$ layer, $H^{(L)}_{r \times n}$; and the abundance matrix $S_{m \times r}$ corresponds to the weights of the output layer, $W^{(L)}_{m \times r}$. This will be satisfied only in the case of the $(L-1)^{th}$ layer having $r$ neurons.

With the identity given in equation 5, it is apparent that the decoder allows no flexibility in the design and the decoder should be strictly in accordance with equation 5 for proper operation of the endmember extraction and the abundance estimation. Nonetheless, when the encoder is considered, the only requirement is that the bottleneck layer should contain $r$ neurons and it may have as many layers before the bottleneck layer as necessary with nonlinear activations, as opposed to the decoder which must be a single layer with all units having linear activations.

*B. Convolutional Autoencoder Architecture*

The CAE consists of multiple convolutional layers and dense layers with a dropout and a flatten layer as shown in the simplified schematic diagram (Fig. 1) of the CAE which was developed with the architecture given in Table I. At its input the 2-D image of each band is fed sequentially to the network.

This monochrome data matrix is convoluted with three convolutional layers which is comprised of $3 \times 3$ convolutional kernels. The first layer has 16 different kernels and the second and the third layers have 8 of those each. Each convolutional layer precedes a max pooling layer of appropriate size in order to generate the respective activation shape for the succeeding convolutional layer as given in Table I. Finally, each valid pixel produced at the end of the convolutional layers are rearranged as a column vector by the flatten layer, which essentially bridges the convolutional layers to the dense layers.

TABLE I. LAYERS OF THE AUTOENCODER

| Layer | Type | Activation Shape |
|---|---|---|
| 1 | Input Layer | $\sqrt{m}, \sqrt{m}, 1$ |
| 2 | Conv2D | $\lceil \sqrt{m}/2 \rceil, \lceil \sqrt{m}/2 \rceil, 16$ |
| 3 | Conv2D | $\lceil \sqrt{m}/4 \rceil, \lceil \sqrt{m}/4 \rceil, 8$ |
| 4 | Conv2D | $\lceil \sqrt{m}/8 \rceil, \lceil \sqrt{m}/8 \rceil, 8$ |
| 5 | Flatten | $\lceil \sqrt{m}/8 \rceil^2 \times 8$ |
| 6 | Dense | $9 \times r$ |
| 7 | Dense | $6 \times r$ |
| 8 | Dropout | $3 \times r$ |
| 9 | Dense | $r$ |
| 10 | Dense | $m$ |

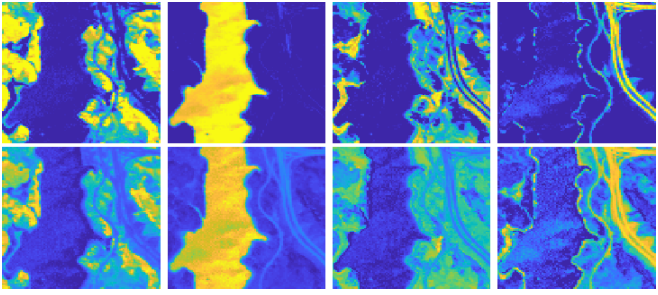

Fig. 2. Abundance maps extracted using CAE in the "Tree", "Water", "Dirt" and "Road" order: Top row - ground truth abundances, Bottom row - extracted abundances using CAE.

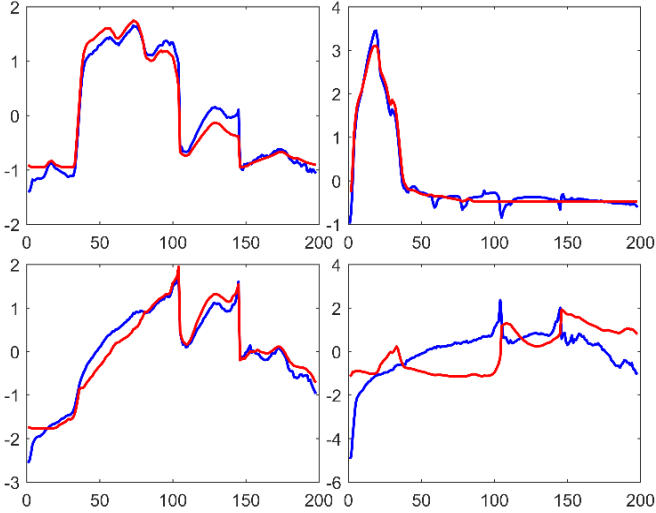

Fig. 3. Endmember spectral signatures (red) extracted using CAE with ground truth (blue): Top left - "Tree", Top right - "Water", Bottom left - "Dirt" and Bottom right - "Road.

Layer 6 and layer 7 have densely connected neurons and the number of neurons are designed according to the predetermined number of endmembers. A dropout layer has been introduced after layer 7 with the dimensions suggested in Table I to avoid the overfitting problem. The dropout[14] rate used was 0.01. Again fully connected layers are used in layer 9 and layer 10 with the number of neurons equal to the number of endmembers and the pixel resolution of the input image, respectively. Activation function for the final layer is linear activation and for the other layers ReLU activation[15] was used. In order to ensure the weights are nonnegative, a nonnegative constraint was added to the output layer. Overfitting of data was intended to avoid with the $\mathbb{L}_2$ regularization with a regularization rate of 0.0001.

To train the autoencoder, MSE was used as the loss function defined as,

$$MSE = \frac{1}{m}\sum_{j=1}^{m}(\hat{x}_{b,j} - x_{b,j})^2 \quad (6)$$

where, $\hat{x}_{b,j}$ is the $j^{th}$ pixel of the reconstructed output and $x_{b,j}$ is the $j^{th}$ pixel of the roll-out input. The NN was trained for 500 epochs using the Adam optimizer.

III. RESULTS AND DISCUSSION

This section is reserved to present the performance of the proposed CAE and to discuss the operation of the architecture using intermediate outputs from the convolutional layers. The first section includes error metrics for the proposed architecture recorded for real hyperspectral data whereas the second section is for the study of the intermediate layers.

A. Performance on Real Hyperspectral Data

In order to evaluate the performance of the proposed algorithm and to assess its competitiveness with other unmixing algorithms, we utilized two performance criteria. Spectral Angle Distance (SAD) is used to evaluate the performance of the endmember extraction process and Root Mean Squared Error (RMSE) is used to evaluate the abundance estimation process of the autoencoder. The SAD and RMSE are given by,

$$RMSE = \sqrt{\frac{1}{m}\sum_{j=1}^{m}(\hat{s}_j - s_j)^2} \quad (7)$$

$$SAD = cos^{-1}\left(\frac{\hat{a}^T a}{\|\hat{a}\|\|a\|}\right) \quad (8)$$

where, $s_j$ and $\hat{s}_j$ are the ground-truth and the estimated abundances for a given endmember, respectively; and $a$ and $\hat{a}$ are the ground-truth endmember and the extracted endmember signature, respectively. For the evaluation of the algorithm, Samson and Jasper-Ridge HSI datasets were used.

Jasper-Ridge is an often tested dataset in the remote sensing literature. Each monochrome image has a pixel resolution of $100 \times 100$ and the dataset is a collection of 224 spectral bands. Yet, only 198 channels can be utilized due to atmospheric and water vapor interference on certain channels. Consequently, the HSI ensemble is $100 \times 100 \times 198$ in size and the ground-truths are given for four endmembers: Dirt, Road, Tree and Water. The experimental results for the Jasper-Ridge dataset are given in Table II and Table III. The estimated abundance maps for the Jasper Ridge dataset are given in Fig. 2 and the extracted endmembers are given in Fig. 3, with respective ground truths.

Another popular dataset is the Samson dataset which is less complex in terms of the spatial distribution of the endmembers. Every single image from those 156 spectral bands has a pixel resolution of $95 \times 95$ and the dataset is already corrected for atmospheric variations. Samson dataset is $95 \times 95 \times 156$ in size and is comprised of three endmembers: Rock, Tree and Water. The experimental results for the Samson dataset are given in Table IV and Table V. For the Jasper-Ridge and the Samson datasets, the proposed CAE has outperformed existing HU algorithms in terms of average RMSE. In addition to this, for the classes where the best performance for RMSE is not from the proposed method, CAE has generated either second best performance or competitive results.

TABLE II. UNMIXING PERFORMANCE FOR JASPER-RIDGE DATASET IN TERMS OF SPECTRAL ANGLE DISTANCE. THE BEST PERFORMANCES ARE IN BOLD TYPEFACE. THE SECOND BEST PERFORMANCES ARE UNDERLINED

| Method | CAE | DNAE | NMF | GNMF | RNMF | VCA |
|---|---|---|---|---|---|---|
| Dirt | **0.1474** | 0.3184 | 1.6357 | 1.0105 | 1.2432 | 1.5352 |
| Road | 1.2473 | **0.4118** | 1.0747 | 1.2817 | 1.7852 | 0.5302 |
| Tree | 0.203 | 0.1263 | **0.1110** | 0.1353 | 0.2264 | 0.4046 |
| Water | 0.1426 | 0.1003 | 0.1413 | 0.1469 | **0.0915** | 0.1387 |
| Average | 0.4351 | **0.2392** | 0.7407 | 0.6436 | 0.8366 | 0.6522 |

TABLE III. UNMIXING PERFORMANCE FOR JASPER-RIDGE DATASET IN TERMS OF ROOT MEAN SQUARE ERROR. THE BEST PERFORMANCES ARE IN BOLD TYPEFACE. THE SECOND BEST PERFORMANCES ARE UNDERLINED.

| Method | CAE | DNAE | NMF | GNMF | RNMF | VCA |
|---|---|---|---|---|---|---|
| Dirt | **0.1926** | 0.2115 | 0.4118 | 0.3388 | 0.2722 | 1.7537 |
| Road | **0.1455** | 0.2054 | 0.2502 | 0.1541 | 0.2047 | 0.2639 |
| Tree | 0.2015 | **0.1991** | 0.2792 | 0.3320 | 0.2630 | 0.3795 |
| Water | 0.1326 | **0.1128** | 0.1575 | 0.1798 | 0.1270 | 2.2116 |
| Average | **0.1681** | 0.1822 | 0.2747 | 0.2512 | 0.2167 | 1.1522 |

TABLE IV. UNMIXING PERFORMANCE FOR SAMSON DATASET IN TERMS OF SPECTRAL ANGLE DISTANCE. THE BEST PERFORMANCES ARE IN BOLD TYPEFACE. THE SECOND BEST PERFORMANCES ARE UNDERLINED.

| Method | CAE | DNAE | NMF | GNMF | RNMF | VCA |
|---|---|---|---|---|---|---|
| Rock | 0.1820 | **0.1189** | 1.7164 | 2.1002 | 0.6281 | 1.8387 |
| Tree | 0.0660 | 0.0467 | **0.0436** | 0.0657 | 0.0488 | 0.0554 |
| Water | 0.7109 | 0.4485 | 0.4951 | 0.6625 | 0.3152 | **0.2004** |
| Average | 0.3196 | **0.2047** | 0.7517 | 0.9428 | 0.3307 | 0.6982 |

TABLE V. UNMIXING PERFORMANCE FOR SAMSON DATASET IN TERMS OF ROOT MEAN SQUARE ERROR. THE BEST PERFORMANCES ARE IN BOLD TYPEFACE. THE SECOND BEST PERFORMANCES ARE UNDERLINED.

| Method | CAE | DNAE | NMF | GNMF | RNMF | VCA |
|---|---|---|---|---|---|---|
| Rock | 0.2837 | **0.2391** | 0.2814 | 0.3962 | 0.5294 | 1.2836 |
| Tree | **0.1814** | 0.3055 | 0.2075 | 0.3921 | 0.3300 | 0.2446 |
| Water | 0.2787 | 0.3152 | **0.2637** | 0.2807 | 0.3312 | 1.1107 |
| Average | **0.2479** | 0.2866 | 0.2509 | 0.3563 | 0.3968 | 0.8796 |

Further, the proposed method has the second best performance in terms of average SAD, for both datasets. Performance of SAD can be improved with smaller Kernel sizes for CNN layers as it allows to recognize spatial features of endmembers with low pixel count. However this increases the computational cost of the CAE. The results obtained using the proposed method were compared with existing methods, such as DNAE, NMF, GNMF, RNMF, and VCA.[8]–[11]. With the introduced change to the autoencoder architecture, the weight parameters of the decoder are associated with the abundances. Hence, the weight optimization of the network, during the training process has allowed the CAE to choose abundances for each pixel, which will minimize the reconstruction error.

*B. The Ablataion Study of the Inner-Working of the Model*

The results from the convolutional layers are given in Fig. 4 for all the channels in each convolutional layer, for a monochrome image. The influence on abundance estimation of convolutional layers is clearly visible. The first convolutional layer generates more detailed images. Though certain channels do not carry information to the succeeding layers, rest of the channels learn the most dominant endmember's properties available in the original monochrome image. At the second and the third convolutional layers lesser original information is available to learn compared to the preceding convolutional layer, hence global abstract patterns tend to approximate to the spatial distribution of the endmember. More dominant a particular endmember is in a band, higher the chances for the global patterns to be the spatial distribution of that endmember. As a result of this convergence the decoder is able to estimate abundances accurately.

## IV. CONCLUSION

In this paper, we have proposed a method for blind HU using CNN, of remotely sensed HSI datasets. The CAE consists of two standard neural architectures: convolutional layer networks and autoencoder architecture. In the autoencoder, the encoder constructs underlying characteristics of the HSI dataset from the feature space produced for each monochrome image by the convolutional filters. This latent space is then used by the decoder to reconstruct the roll-out image of the monochrome image which is given as the input to the CAE as the output of the complete architecture. When the network is thoroughly trained, the weights of the output layer provides the abundances; and the activations from the layer before the output layer provides the endmember spectral signatures, simultaneously. Experimental results were produced for two real HSI datasets: Jasper-Ridge and Samson. Error metrics recorded for the two datasets indicate that the proposed method is successful in estimating the abundances for a given HSI dataset. Moreover, it is evident that the proposed CAE outperforms existing HU methods in terms of abundance estimation. This was achieved with the proposed change for the autoencoder architecture, which performs the optimization on the abundance maps unlike other DL based HU algorithms found in the literature, where the conventional approach is to optimize the endmember signatures. Furthermore, the convolutional layers improve the abundance estimation by recognizing the spatial distribution of the endmembers.

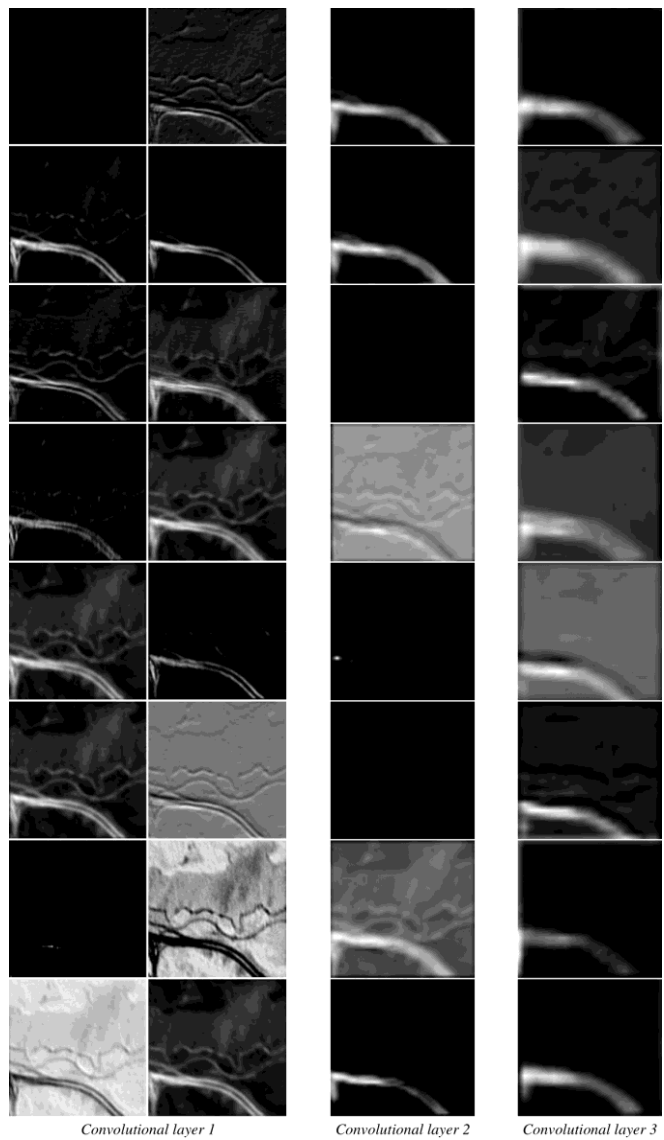

Fig. 4. Intermediate output of the convolutional layers.